\documentclass{article}

\usepackage{graphicx}
\usepackage{amsmath}

\title{Predicting growth fluctuation in network economy}
\author{Yoshiharu Maeno}
\begin{document}

\maketitle

\begin{abstract}
This study presents a method to predict the growth fluctuation of firms interdependent in a network economy. The risk of downward growth fluctuation of firms is calculated from the statistics on Japanese industry.
\end{abstract}

\section{Introduction}

Does an abrupt ill health of one firm have a big impact on the health of others? 

Many firms ended in disastrous failure during the worldwide financial crisis in 2008. Since then, risk managers, executives, and investors have been strongly interested in the transmission of distress and the knock-on defaults between firms which are interdependent in a network economy \cite{May10}. In this study, a model for such firms is formulated with stochastic differential equations. Probability parameters on trades between firms can be inferred statistically, and the time evolution of the net-worth of the firms can be predicted. The conditional value at risk of the downward growth fluctuation of firms is calculated from the statistics on Japanese industry in 2005.

\section{Stochastic model}

A stochastic model for firms is presented. Time dependent variables $a_{i}(t)$ for $i=1,2,\cdots,N$ is the net-worth of the $i$-th firm at time $t$. The interplay between the firms governs the time evolution of $a_{i}(t)$. The income by the sales to others lets $a_{i}(t)$ increase. Its rate of change is given by $\phi_{ij} a_{j}(t)$ where the probability parameters $\phi_{ij}$ are constant. The expenditure on the purchases from others and labor wages lets $a_{i}(t)$ decrease. The rate is $-\lambda_{i} a_{i}(t)$ where $\lambda_{i}$ is constant. This is a special case (linear production function) of the model for the financial accelerator in credit networks \cite{Del10}. It is also similar to the model for evolutionary autocatalytic sets \cite{Meh09}.

The stochasticity of the time evolution ensues from an unpredictably irregular pattern of trades between firms. The number of trades obeys a Poisson distribution if the probability of a trade per unit time is constant. The amplitude of fluctuation is nealy the square root of the average. The time evolution of $a_{i}(t)$ is given by a system of stochastic differential equations in eq.(\ref{dait/dt}). The functional form of the Gaussian white noises $\xi^{{\rm [I]}}_{j}(t)$ and $\xi^{{\rm [E]}}_{i}(t)$ is not known.
\begin{equation}
\frac{{\rm d}a_{i}(t)}{{\rm d}t} = \sum_{j=1}^{N} \phi_{ij} a_{j}(t) + \sum_{j=1}^{N} \sqrt{\phi_{ij} a_{j}(t)}\xi^{{\rm [I]}}_{j}(t) - \lambda_{i} a_{i}(t) - \sqrt{\lambda_{i} a_{i}(t)} \xi^{{\rm [E]}}_{i}(t).
\label{dait/dt}
\end{equation}

Eq.(\ref{dait/dt}) is equivalent to the Fokker-Planck equation in eq.(\ref{FokkerPlanck}). It is a partial differential equation, which describes the time evolution of the joint probability density function $P(\mbox{\boldmath{$a$}},t)$ of probability variables $a_{i}$ at $t$.
\begin{equation}
\frac{\partial P(\mbox{\boldmath{$a$}},t)}{\partial t} = -\sum_{i=1}^{N} \frac{\partial}{\partial a_{i}} A_{i}(\mbox{\boldmath{$a$}}) P(\mbox{\boldmath{$a$}},t) + \frac{1}{2} \sum_{i,j=1}^{N} \frac{\partial^{2}}{\partial a_{i} \partial a_{j}} B_{ij}(\mbox{\boldmath{$a$}}) P(\mbox{\boldmath{$a$}},t).
\label{FokkerPlanck}
\end{equation}

The drift and diffusion coefficients in eq.(\ref{FokkerPlanck}) are given by eq.(\ref{tildeA}) and (\ref{tildeB}). 
\begin{equation}
A_{i}(\mbox{\boldmath{$a$}}) = \sum_{j=1}^{N} \tilde{A}_{ij} a_{j} = \sum_{j=1}^{N} (\phi_{ij}-\lambda_{i}\delta_{ij})a_{j}.
\label{tildeA}
\end{equation}
\begin{equation}
B_{ij}(\mbox{\boldmath{$a$}}) = \sum_{k=1}^{N} \tilde{B}_{ijk} a_{k} = \sum_{k=1}^{N} \{(\phi_{ik}+\lambda_{i}\delta_{ik})\delta_{ij}+\sqrt{\phi_{ik}\phi_{jk}}(1-\delta_{ij})\} a_{k}.
\label{tildeB}
\end{equation}

Predicting $a_{i}(t)$, given $\mbox{\boldmath{$\phi$}}$ and $\mbox{\boldmath{$\lambda$}}$, is a forward problem. Inferring the value of $\mbox{\boldmath{$\phi$}}$ and $\mbox{\boldmath{$\lambda$}}$ from the observation on $a_{i}(t)$ statistically is an inverse problem \cite{Mae10}. These problems are mixed under practical conditions. The values of some parameters are known, and some data on $a_{i}(t)$ are given. Eq.(\ref{FokkerPlanck}) is converted to a system of ordinary differential equations in eq.(\ref{dm/dt}) and (\ref{dv/dt}), which describe the time evolution of the 1st and 2nd order moments $\mu^{[1]}_{i}(t)$ and $\mu^{[2]}_{ij}(t)$.
\begin{equation}
\frac{{\rm d} \mu^{[1]}_{i}(t)}{{\rm d}t} = \sum_{j=1}^{N} \tilde{A}_{ij} \mu^{[1]}_{j}(t). 
\label{dm/dt}
\end{equation}
\begin{equation}
\frac{{\rm d} \mu^{[2]}_{ij}(t)}{{\rm d}t} = \sum_{k=1}^{N} \tilde{A}_{ik} \mu^{[2]}_{kj}(t) + \tilde{A}_{jk} \mu^{[2]}_{ki}(t) + \tilde{B}_{ijk} \mu^{[1]}_{k}(t). 
\label{dv/dt}
\end{equation}

Generally, the time evolution of the $m$-th order moments is given by a system of linear differential equations in eq.(\ref{dmu/dt}). The elements of the $N^{m} \times 1$ vector $\mbox{\boldmath{$\mu$}}^{[m]}(t)$ are $\mu^{[m]}_{11\cdots1}(t),\ \mu^{[m]}_{11\cdots2}(t),\ \cdots,\ \mu^{[m]}_{NN \cdots N}(t)$. The $N^{m} \times N^{m}$ matrix $\mbox{\boldmath{$A$}}^{[m]}$, and $N^{m} \times N^{m-1}$ matrix $\mbox{\boldmath{$B$}}^{[m]}$ are calculated from $\tilde{A}_{ij}$ and $\tilde{B}_{ijk}$.
\begin{equation}
\frac{{\rm d} \mbox{\boldmath{$\mu$}}^{[m]}(t)}{{\rm d} t} = \mbox{\boldmath{$A$}}^{[m]} \mbox{\boldmath{$\mu$}}^{[m]}(t) + \mbox{\boldmath{$B$}}^{[m]} \mbox{\boldmath{$\mu$}}^{[m-1]}(t).
\label{dmu/dt}
\end{equation}

The solution of eq.(\ref{dmu/dt}) with the initial conditions $\mbox{\boldmath{$\mu$}}^{[m]}(0)$ is given by eq.(\ref{sol}). 
\begin{equation}
\mbox{\boldmath{$\mu$}}^{[m]}(t) = \exp(\mbox{\boldmath{$A$}}^{[m]} t) (\int_{0}^{t} \exp(-\mbox{\boldmath{$A$}}^{[m]} t') \mbox{\boldmath{$B$}}^{[m]} \mbox{\boldmath{$\mu$}}^{[m-1]}(t') {\rm d}t' + \mbox{\boldmath{$\mu$}}^{[m]}(0)).
\label{sol}
\end{equation}

Approximately, $P(\mbox{\boldmath{$a$}},t)$ is a multi-variate normal distribution with the mean $\mu^{[1]}_{i}(t)$ and covariance $\mu^{[2]}_{ij}(t)$. The exact formula for $P(\mbox{\boldmath{$a$}},t)$ is obtained by the Edgeworth series. It is an asymptotic expansion of $P(\mbox{\boldmath{$a$}},t)$ in terms of cumulants. The logarithmic likelihood function $L(\mbox{\boldmath{$\phi$}},\mbox{\boldmath{$\lambda$}})$ is obtained immediately, once eq.(\ref{FokkerPlanck}) is solved, given a dataset $\mbox{\boldmath{$a$}}^{{\rm [D]}}(t_{d})$ at $t_{d}$ for the observations $d=1,2,\cdots$. It is given by eq.(\ref{lik}). The estimators $\hat{\mbox{\boldmath{$\phi$}}}$ and $\hat{\mbox{\boldmath{$\lambda$}}}$ are those which maximize $L$.
\begin{equation}
L(\mbox{\boldmath{$\phi$}},\mbox{\boldmath{$\lambda$}}) = \sum_{d} \log (P(\mbox{\boldmath{$a$}}^{{\rm [D]}}(t_{d}),t_{d}|\mbox{\boldmath{$\phi$}},\mbox{\boldmath{$\lambda$}})).
\label{lik}
\end{equation}

\section{Growth fluctuation}

\begin{figure}
\begin{center}
\includegraphics[scale=0.4,angle=-90]{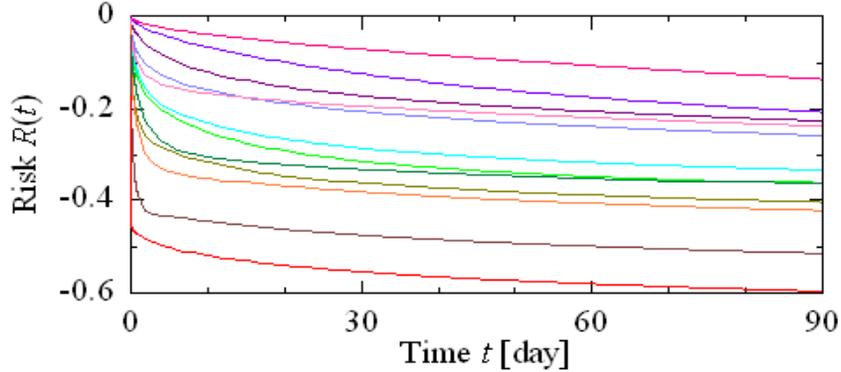}
\end{center}
\caption{Risk $R_{i|j}^{[0.01]}(t)$ of the representative firms as a function of $t$. The $j$-th sector is Transportation equipment. The $i$-th sector is one of 12 selected sectors.}
\label{021801s}
\end{figure}

The risk of firms are defined, and calculated with the Leontief coefficients in the input-output model for $N=34$ Japanese industry sectors in 2005\footnote{Ministry of Internal Affairs and Communications, 2005 Input-Output Tables for Japan. http://www.stat.go.jp/english/data/io/io05.htm.}. The $q$ quantile value at risk $V_{j}^{[q]}(t)$ of the $j$-th firm at $t$ is defined by eq.(\ref{Vardef}) \cite{Mar10}. It is the net-worth at which the cumulative density is $q$. $P_{{\rm M}}(a_{j},t)$ is the marginal probability density function of $a_{j}$.
\begin{equation}
\int_{-\infty}^{V_{j}^{[q]}(t)} P_{{\rm M}}(a_{j},t) {\rm d} a_{j} = q.
\label{Vardef}
\end{equation}

The conditional value at risk $C_{i|j}^{[q]}(t)$ which the ill health of the $j$-th firm imposes on the $i$-th firm at $t$ is defined by eq.(\ref{CVardef}). $P_{{\rm C}}(a_{i},t|a_{j})$ is the probability density function of $a_{i}$ conditioned on the value of $a_{j}$. Note $C_{i|j}^{[q]}(t) \neq C_{j|i}^{[q]}(t)$.
\begin{equation}
\int_{-\infty}^{C_{i|j}^{[q]}(t)} P_{{\rm C}}(a_{i},t|V_{j}^{[q]}(t)) {\rm d} a_{i} = q.
\label{CVardef}
\end{equation}

The quantity $R_{i|j}^{[q]}(t)$ is the risk of downward fluctuation from the expected growth at $t$. If $P(\mbox{\boldmath{$a$}},t)$ is a multi-variate normal distribution, $R_{i|j}^{[q]}(t)$ is given by eq.(\ref{Ri|j}) where $\psi^{[q]} = {\rm erf}^{-1}(2q-1)$. For example, $\psi^{[0.05]}=-1.16$ for 5 percentile and $\psi^{[0.01]}=-1.65$ for 1 percentile.
\begin{equation}
R_{i|j}^{[q]}(t) = \frac{ C_{i|j}^{[q]}(t) - \mu^{[1]}_{i}(t) }{ \mu^{[1]}_{i}(t)} = \frac{ \sqrt{2 \frac{\mu^{[2]}_{ij}(t)^{2}}{\mu^{[2]}_{jj}(t)}} + \sqrt{2( \mu^{[2]}_{ii}(t) - \frac{\mu^{[2]}_{ij}(t)^{2}}{\mu^{[2]}_{jj}(t)})} }{\mu^{[1]}_{i}(t)} \psi^{[q]}.
\label{Ri|j}
\end{equation}

The Leontief coefficients determine $\mbox{\boldmath{$\phi$}}$. The past growth rates of the sectors are used to obtain $\hat{\mbox{\boldmath{$\lambda$}}}$. Suppose a representative firm in each industry sector whose share of production is $1\%$. Figure \ref{021801s} shows $R^{[0.01]}_{i|j}(t)$ of the firms in 12 selected sectors ($i$) as a function of $t$. The $j$-th sector is Transportation equipment. The risk increases as time goes by. The representative firm in Mining have the largest risk, $R \approx -0.6$ at a quarter later ($t=90$ days), when the representative firm in Transportation equipment falls ill. It is followed by the firms in Office supplies, Textile products, Non-ferrous metals, Electronic parts, Finance and insurance, Precision instruments, Iron and steel, Chemical products, Commerce, and Public administration. The firm in Medical service, health, social security and nursing care has the smallest risk.

\end{document}